\documentclass{article}

\usepackage{arxiv}

\usepackage[utf8]{inputenc} %
\usepackage[T1]{fontenc}    %
\usepackage{hyperref}       %
\usepackage{url}            %
\usepackage{booktabs}       %
\usepackage{amsfonts}       %
\usepackage{nicefrac}       %
\usepackage{microtype}      %
\usepackage{graphicx}
\usepackage{float}
\usepackage{xcolor}
\usepackage{caption}
\usepackage{subcaption}
\graphicspath{ {figures/} }

\title{Continental-scale building detection \\from high resolution satellite imagery}

\author{
Wojciech Sirko, Sergii Kashubin, Marvin Ritter, Abigail Annkah, Yasser Salah Eddine Bouchareb, \\
\bf{Yann Dauphin, Daniel Keysers, Maxim Neumann, Moustapha Cisse, John Quinn\thanks{Address for correspondence: \texttt{open-buildings-dataset@google.com}}} \\
Google Research
}

\begin{document}
\maketitle

\begin{abstract}
Identifying the locations and footprints of buildings is vital for many practical and scientific purposes. Such information can be particularly useful in developing regions where alternative data sources may be scarce. In this work, we describe a model training pipeline for detecting buildings across the entire continent of Africa, using 50 cm satellite imagery. Starting with the U-Net model, widely used in satellite image analysis, we study variations in architecture, loss functions, regularization, pre-training, self-training and post-processing that increase instance segmentation performance. Experiments were carried out using a dataset of 100k satellite images across Africa containing 1.75M manually labelled building instances, and further datasets for pre-training and self-training. We report novel methods for improving performance of building detection with this type of model, including the use of mixup (mAP +0.12) and self-training with soft KL loss (mAP +0.06). The resulting pipeline obtains good results even on a wide variety of challenging rural and urban contexts, and was used to create the Open Buildings dataset of 516M Africa-wide detected footprints. 
\end{abstract}

\section{Introduction}

Building footprints are useful for a range of important applications, from mapping, population estimation and urban planning to humanitarian response and environmental science. In developing regions this information can be particularly valuable, for instance in areas with infrequent censuses, or with a high prevalence of informal settlements, or where there is rapid change, such as in emerging mega-cities. Although the detection of buildings in developing regions can be technically challenging, it has the potential to address large information gaps with respect to current knowledge.

\begin{figure}[tbp]
    \centering
    \begin{subfigure}{0.16\textwidth}
        \centering
        \includegraphics[width=\linewidth]{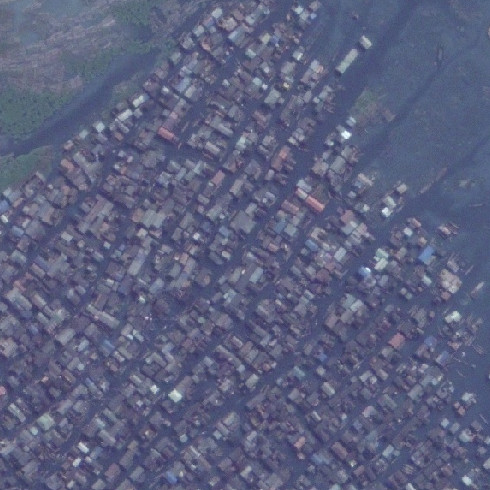} 
        \caption{}
    \end{subfigure}
    \begin{subfigure}{0.16\textwidth}
        \centering
        \includegraphics[width=\linewidth]{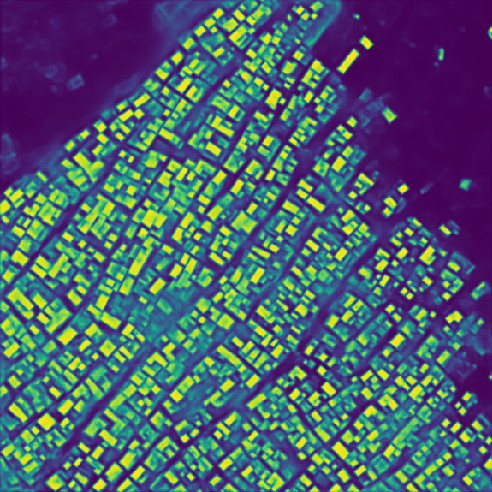} 
        \caption{}
    \end{subfigure}
    \begin{subfigure}{0.16\textwidth}
        \centering
        \includegraphics[width=\linewidth]{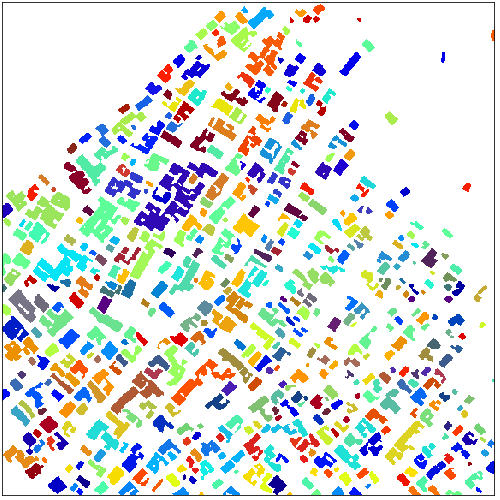} 
        \caption{}
    \end{subfigure}
    \hfill
    \begin{subfigure}{0.16\textwidth}
        \centering
        \includegraphics[width=\linewidth]{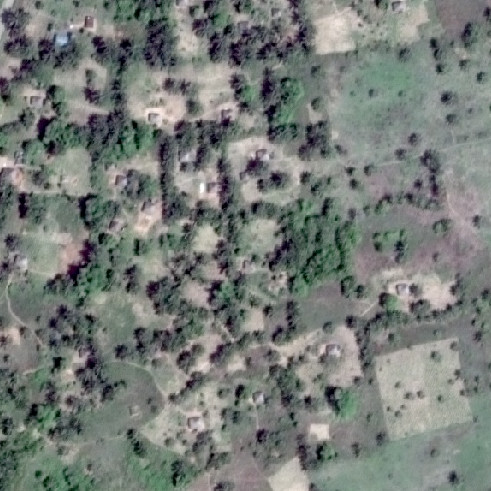} 
        \caption{}
    \end{subfigure}
    \begin{subfigure}{0.16\textwidth}
        \centering
        \includegraphics[width=\linewidth]{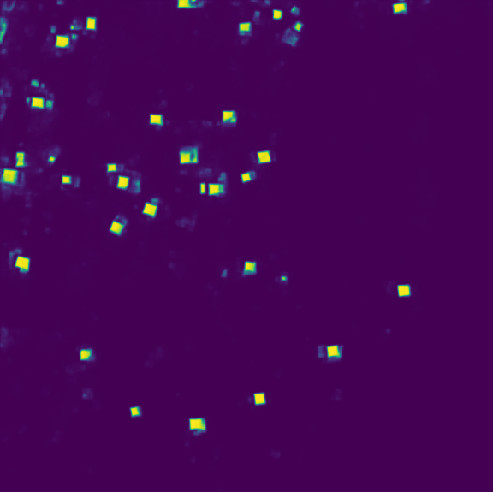} 
        \caption{}
    \end{subfigure}
    \begin{subfigure}{0.16\textwidth}
        \centering
        \includegraphics[width=\linewidth]{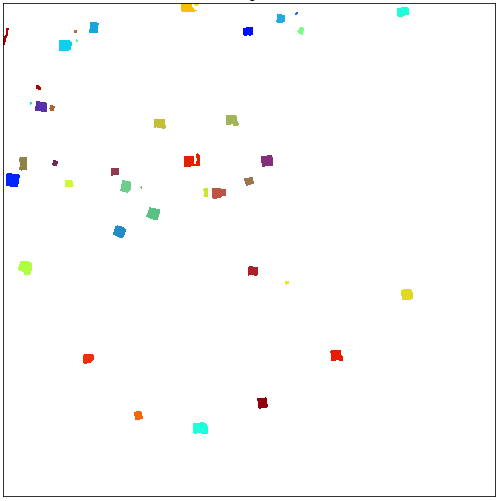} 
        \caption{}
    \end{subfigure}
    \caption{Examples of bottom-up building detection: (a,d) test images; (b,e) semantic segmentation confidences; (c,f) instances of buildings found. Satellite imagery in this paper: Maxar Technologies, CNES/Airbus.}
\label{fig:segmentation_examples}
\end{figure}

In this work, we describe the development of a detection pipeline for identifying building footprints across the continent of Africa from satellite imagery of 50 cm resolution. The land surface of Africa is about 20\% of the Earth's total and has a wide diversity of terrain and building types, meaning that this is a broad and challenging problem. Challenges include the range of geological or vegetation features which can be confused with built structures, settlements with many contiguous buildings not having clear delineations, and areas characterised by small buildings, which can appear only a few pixels wide at this resolution. In rural or desert areas, buildings constructed with natural materials can visually blend in to the surrounding area. Figures~\ref{fig:segmentation_examples} and \ref{fig:labelling_policy} show some examples.

Progress in deep learning methods with remote sensing imagery has created new possibilities for working at this scale, and recent work on building detection from high resolution satellite imagery has shown remarkable improvements in precision and recall, which we review briefly in Section~\ref{sec:related}. Common to much of this work is the U-Net architecture~\cite{unet}, an encoder-decoder model for semantic segmentation. Rather than learning to identify building instances with an end-to-end model, the idea in this type of `bottom-up' segmentation is to classify each pixel of an aerial image as \textit{building} or \textit{non-building}, and then to find connected components at some confidence score threshold. Illustrations of our model operating in this way are shown in Figure~\ref{fig:segmentation_examples}. Existing studies have tended to be limited to particular cities or countries, however, leaving an open question as to how well such methods generalise to wider areas, particularly in developing regions.

We begin by describing training and evaluation datasets compiled for this work in Section~\ref{sec:datasets}, including weakly labelled and unlabelled image data for pre-training and self-training respectively. We then describe a number of methods tested to improve building detection performance, in the following categories:

\begin{itemize}
    \item Choices of \textbf{architecture}, concerning different encoders and decoders (Section~\ref{sec:model}).
    \item \textbf{Loss functions} which are more appropriate for building segmentation than generic segmentation choices (Section~\ref{sec:losses}).
    \item \textbf{Regularization}, including mixup and other augmentations (Section~\ref{sec:regularization}).
    \item \textbf{Pre-training} (Section~\ref{sec:pretraining}).
    \item \textbf{Self-training} methods for improving building detection performance using additional unlabelled data (Section~\ref{sec:selftraining}).
    \item \textbf{Pre-processing} methods for preparing the input image and labels, including morphological adjustments (Section~\ref{sec:preprocessing}).
    \item \textbf{Post-processing} methods for converting semantic segmentation predictions into predicted instances (Section~\ref{sec:postprocessing}).
\end{itemize}

We report experimental results in Section~\ref{sec:evaluation}, including ablation studies to determine the effectiveness of different methods, and evaluation of the accuracy and consistency of the resulting building detection pipeline in different contexts. 

In summary, the main contributions of this work are: (1) we provide the first experimental results, to our knowledge, on the training and evaluation of building detection models on high resolution aerial imagery at a continental scale, (2) we propose a number of specific methods for improving building detection performance using the U-Net model, including mixup, self-training, distance weighting with Gaussian convolutions, and residual decoder blocks, and (3) the resulting pipeline was used to generate an open dataset of 516M building footprints across Africa, available at \url{https://sites.research.google/open-buildings}.

\section{Related work}
\label{sec:related}

Instance segmentation is a well-studied task, though most literature on instance segmentation methods are concerned with detecting objects in photos. In such settings, the best performing methods tend to be end-to-end instance segmentation models, which break down the problem into feature extraction, bounding box regression and mask prediction stages; recent examples are YOLOv4 \cite{wang2020scaled} or Hybrid Task Cascades \cite{chen2019hybrid}. Satellite imagery, however, has different characteristics, which has motivated alternative approaches to instance segmentation. In particular, objects such as buildings can be smaller and more densely clustered, which is a challenge for methods that use bounding box regression with a non-maximum suppression (NMS) step, since in cases where instances are densely arranged, NMS can suppress true detections and reduce recall.

In satellite imagery, therefore, a more common approach is to first carry out semantic segmentation to classify each pixel in an image as \textit{building} or \textit{non-building}. Post-processing is then done to extract instances, for example by thresholding and finding connected components. An example of this type of encoder-decoder approach that has been successful for building detection is TernausNetV2 \cite{iglovikov2018ternausnetv2}, which uses U-Net \cite{unet} with a ResNet-based encoder, and three output classes---building, non-building, and touching edges---in order to emphasise the boundary regions between instances. Other successful building detection methods have used different ways of increasing the model's focus on edges of nearby instances, such as distance-based weighting of pixels in the loss \cite{neptuneai18opensolution}.

The CVPR DeepGlobe Challenge \cite{demir2018deepglobe} posed three satellite imagery tasks: building detection, road detection and land cover mapping. Of the 22 top entries for building detection, 13 were based on U-Net~\cite{unet} and only one used an end-to-end instance segmentation model (Mask-RCNN~\cite{maskrcnn}). The SpaceNet challenge \cite{van2018spacenet} has convened a number of building detection challenges, most recently the Multi-Temporal Urban Development Challenge \cite{zhang2020multi}, for which four of the top five entries were based on U-Net. The overall best performing method was HRNet \cite{zhang2020multi}, a semantic segmentation model with a different architecture to U-Net, in that it dispenses with a decoder stage and uses adaptive spatial pooling.

While progress has been made on methods for building detection in satellite imagery, the available evidence in the literature and from competitions is limited in geographical scope. The SpaceNet buildings dataset covers six cities: Atlanta, Khartoum, Las Vegas, Paris, Rio de Janeiro, and Shanghai. The SpaceNet Multi-Temporal Urban Development dataset contains labelled images from much more diverse geography (41,000km$^2$ of imagery in 101 locations), although given the nature of the challenge, the locations are mainly semi-urban. Image resolution in this dataset is 4m per pixel, which also means that detections are limited to larger buildings. In this work, we provide the first empirical results on the feasibility of detecting the majority of buildings across an entire continent from 50 cm imagery, assessing model generalisation across many types of terrain and cultures/styles in widely differing urban and rural settings.

\section{Datasets}
\label{sec:datasets}

We next describe the continent-wide datasets prepared for the training and evaluation of building detection models, and with varying levels of labelling. The first category is a set of satellite images with full instance labels, used for conventional supervised learning and also the basis of our evaluation. Secondly, we prepared a larger set of images with class labels corresponding to pretext tasks, suitable for pre-training. Thirdly, we prepared a set of images with no labels at all, used for unsupervised self-training. For additional evaluation of the final dataset, we also prepared a sparsely-labelled evaluation dataset. These datasets are summarised in Table~\ref{tab:datasets}.

\begin{table}[tbp]
\caption{Summary of datasets prepared for this work. All images are 600$\times$600 pixels, at 50 cm resolution.}
\centering
\begin{tabular}{llll}
\toprule
Type/Usage & Number of images & Labels & Number of instances \\
\midrule
Training & 99,902 & Building instances & 1.67M \vspace{.2cm}\\
Evaluation & 1,920 & Building instances & 80,672 \vspace{.2cm} \\
Pre-training & 1M & \begin{tabular}[c]{@{}l@{}}Coarse location,\\Fine location,\\ Nighttime luminance\end{tabular} & - \vspace{.2cm} \\
Self-training & 8.7M & - & - \vspace{.2cm} \\  
Additional evaluation & 0.9M & Sparse building instances & 0.9M \\  
\bottomrule
\end{tabular}
\label{tab:datasets}
\end{table}

\subsection{Supervised learning and evaluation data}
\label{sec:fully-labelled-data}

\begin{figure}[tbp]
    \centering
    \begin{subfigure}[t]{0.2\textwidth}
        \centering
        \includegraphics[width=\linewidth]{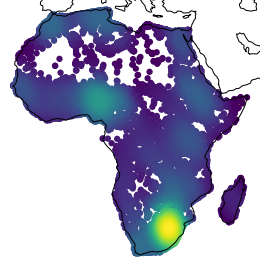} 
        \caption{}
    \end{subfigure}
    \begin{subfigure}[t]{0.2\textwidth}
        \centering
        \includegraphics[width=\linewidth]{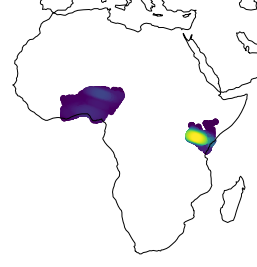} 
        \caption{}
    \end{subfigure}
    \begin{subfigure}[t]{0.2\textwidth}
        \centering
        \includegraphics[width=\linewidth]{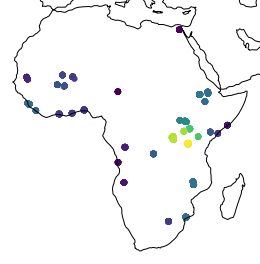} 
        \caption{}
    \end{subfigure}
    \begin{subfigure}[t]{0.2\textwidth}
        \centering
        \includegraphics[width=\linewidth]{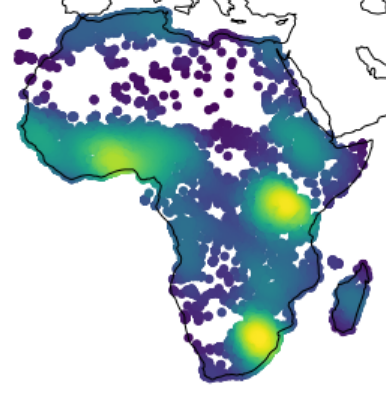} 
        \caption{}
    \end{subfigure}
    \caption{Geographical distribution of data: (a) training set component with 18,149 images and 120,236 building polygons; (b) training set component with 81,753 images and 1.55M building polygons; (c) test set with 1,920 images and 80,672 building polygons; (d) additional sparse evaluation data with 0.9M images and 0.9M building polygons.}
\label{fig:data_geographical_distribution}
\end{figure}

We collected a training set of 99,902 RGB satellite images of size 600$\times$600 pixels, of locations across the African continent. Figure~\ref{fig:data_geographical_distribution} shows the geographical distribution of these images. Given data resources available to us, these were composed of two different sets with different geographical densities. The resulting training set has broad coverage across the continent, with particular concentrations of images for locations in East and West Africa.

Test locations were chosen according to more specific criteria. When sampling random locations across large areas, most images do not contain any buildings. In order to avoid having an evaluation set which was biased towards rural and empty areas, a set of 47 specific regions of interest was selected. These were chosen to contain a mix of rural, medium-density 
and urban areas in different regions of the continent, including informal settlements in urban areas as well as refugee facilities.

\begin{figure}
    \centering
    \includegraphics[width=\textwidth]{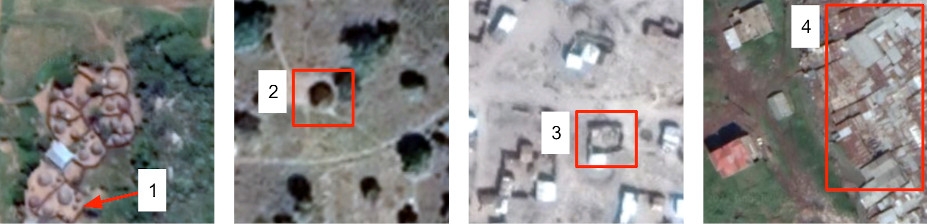}
    \caption{Examples of building labelling policy, taking into account characteristics of different areas across the African continent.  (1) Example of a compound containing both dwelling places as well as smaller outbuildings such as grain stores: the smaller buildings should be ignored. (2) Example of a round, thatched-roof structure which is difficult to distinguish from trees: use cues about pathways, clearings and shadows to disambiguate. (3) Example of internal shadow, indicating that this is an enclosure wall and not a building. (4) Example of several contiguous buildings for which the boundaries cannot be distinguished, where the `dense buildings' class should be used.
    }
    \label{fig:labelling_policy}
\end{figure}

The labelling policy was developed to take into account characteristic settings across the continent, with some examples shown in Figure~\ref{fig:labelling_policy}. One challenge is the labelling of small buildings, as structures a few metres across can be close to the limit of detectability in 50 cm imagery. Another challenge is the labelling of buildings which are densely positioned in close proximity to each other. We introduced a \emph{dense building} class for labelling, when a human annotator was not able to ascertain the exact boundary between individual buildings. This is analogous to the \emph{crowd} type in COCO \cite{caesar2018coco}.  

\subsection{Pre-training data}
\label{sec:pretrain-datasets}

We generated further datasets of satellite imagery, with classification labels for alternative tasks which were used as the basis for representation learning experiments and pre-training. A convenient feature of satellite imagery is that every pixel is associated with a longitude and latitude, so that it can be linked to various other geospatial data. For example, Jean et al. \cite{jean2016combining} demonstrated the use of nighttime lights data to be the basis of a pretext task, such that a model trained to predict how bright a location is at night from daytime imagery learns a representation of satellite imagery which helps as a starting point for other tasks.

We sampled one million images of size 600$\times$600 pixels, at 50 cm per pixel resolution from across the continent of Africa. Sampling density was not completely uniform, as source imagery was limited e.g.\ within large deserts and other uninhabited areas.

\begin{figure}
    \centering
    \includegraphics[width=.495\textwidth]{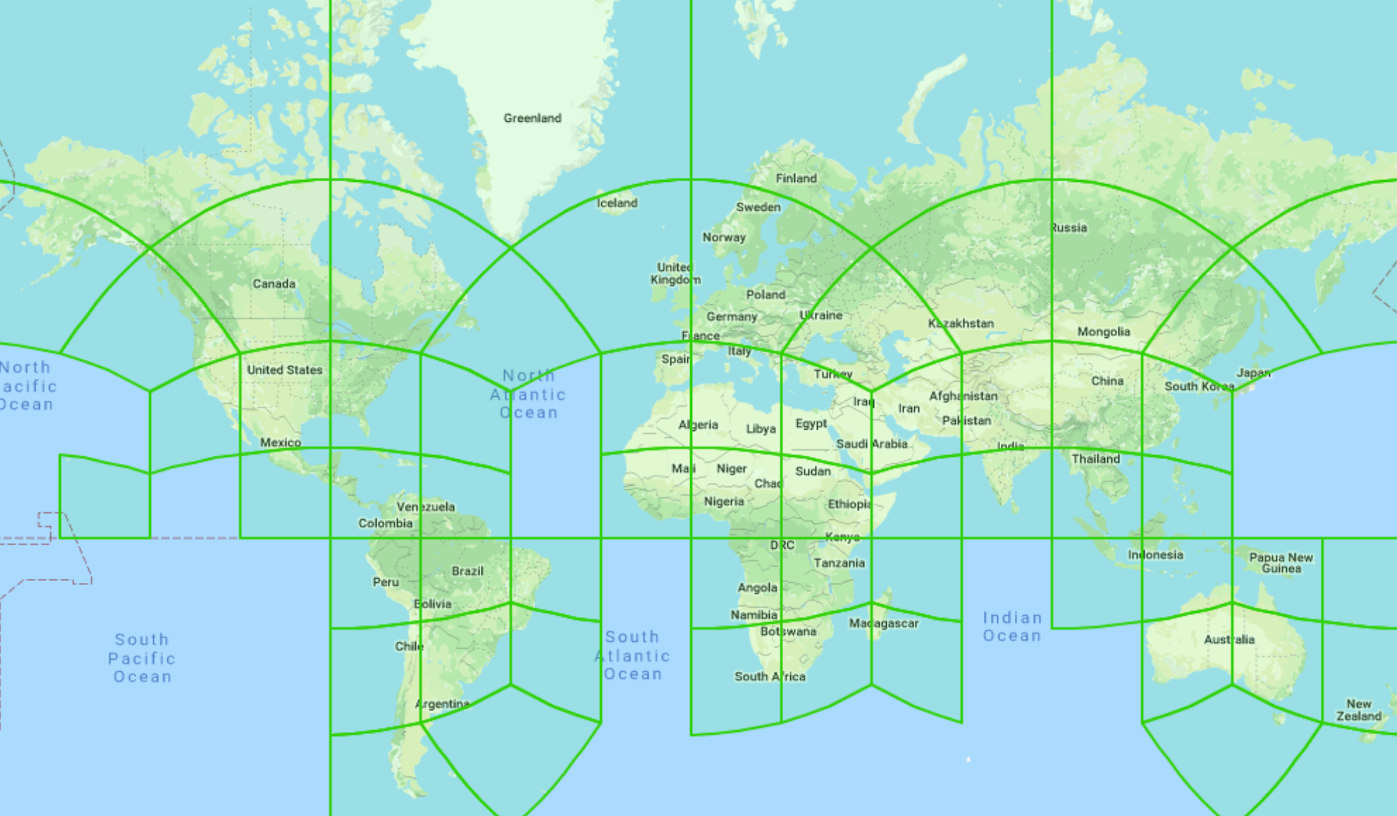}
    \includegraphics[width=.495\textwidth]{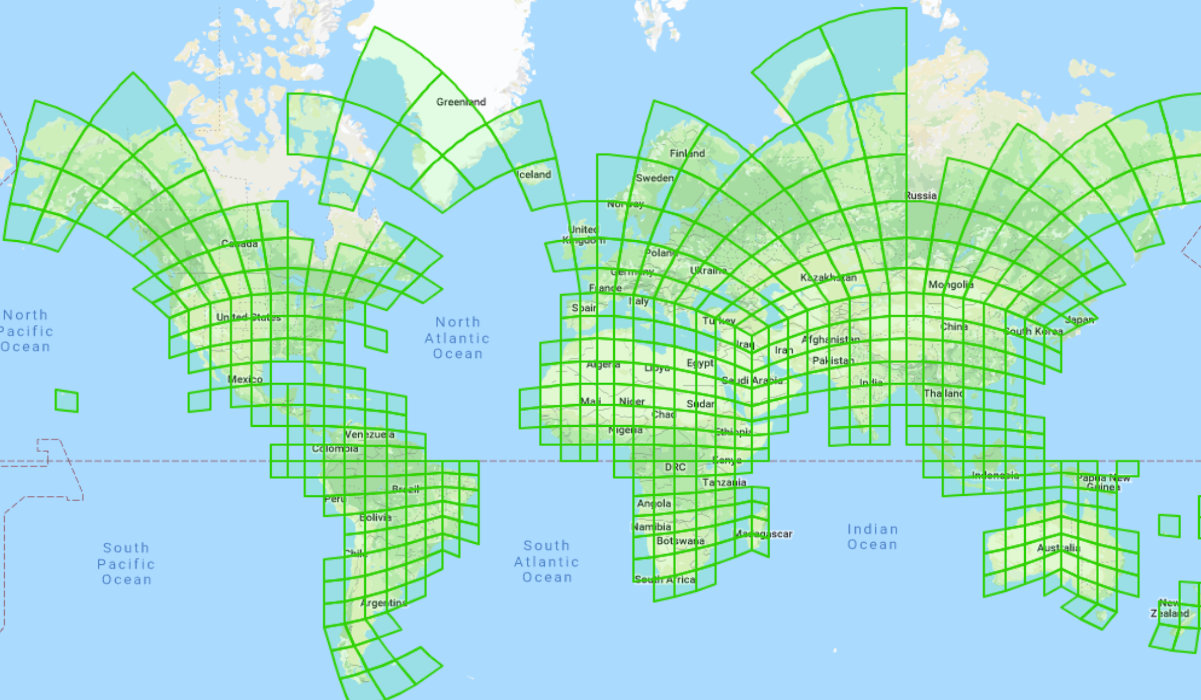}
    \caption{Partitioning of landmass into cells of roughly equal area, according to S2 geometry: coarse (left) and fine (right).}
    \label{fig:s2_areas}
\end{figure}

For each of these images, we computed information which could be used as pretext task labels. We used the location of the image as a classification target, by binning the Earth's surface into parcels of roughly equal area based on S2 geometry\footnote{https://s2geometry.io/}, as shown in Figure~\ref{fig:s2_areas}. This gives us a classification task, in which the goal is to predict for an image patch which part of the world it comes from. The intuition is that in order to obtain good performance on this task, a model might learn to distinguish different vegetation, architectural or geographical features. We also computed nighttime lights data, using DMSP OLS sensor data. This data is computed by averaging nighttime light luminance over the course of a year, in order to correct for temporal factors such as cloud cover. Following the methodology in \cite{xie2016transfer}, we binned the luminance values into four classes, and also retained the original values. Using this as a supervision label predisposes the model to pay attention to human-constructed features such as buildings, which emit light. The methods used to create pre-trained checkpoints with these datasets are described in Section~\ref{sec:pretraining}.

\subsection{Self-training data}
\label{subsec:self-training-dataset}

This unlabeled dataset was created by sampling 100M 640$\times$640 pixel satellite images from the African continent. More than 90\% of images contained no buildings, therefore we subsampled the dataset using our best supervised model, so that only around $\frac{1}{8}$ of images did not contain buildings. The final dataset after filtering contained 8.7M images.

\subsection{Additional evaluation data}
\label{subsec:additional-eval-dataset}

This sparsely labeled dataset contains 0.9M 448$\times$448 pixel satellite images from the African continent and is a by-product of the internal Google Maps evaluation process. Each image is centered on one building detection (not necessarily from our model), and therefore contains a mixture of images with buildings and images with features that are easily confused as buildings, such as rocks or vegetation. For each image, a human evaluator assessed whether that central point contains a building. If so, they created a label with the footprint of that single building, and if not the label is empty. Around $\frac{1}{8}$ of the images in this dataset were centered on non-buildings. This dataset can therefore be used for estimating precision, but not recall. It has good coverage of the African continent, but due to the sampling process, the density of images does not match the real building density in all locations. See Section~\ref{sec:evaluation} for how we used this dataset.

\section{Model}
\label{sec:model}

Our experiments are based on the U-Net model \cite{unet}, which is commonly used for segmentation of satellite images. As this is a semantic segmentation model, we use it to classify each pixel in an input image as \textit{building} or \textit{non-building}. To convert this to an instance segmentation, we threshold the predictions at some confidence level, and search for connected components (shown in Figure~\ref{fig:segmentation_examples}, where we convert from pixel-wise confidences in panels (b) and (e) to detected instances in panels (c) and (f)).

U-Net is an encoder-decoder architecture, and we use an encoder based on ResNet-50-v2~\cite{resnetv2}. Preliminary experiments with ResNet-v2-101 and ResNet-v2-152 suggested that deeper encoder architectures did not improve accuracy.

\paragraph{Residual decoder}

\begin{figure}
    \centering
\begin{subfigure}[t]{0.33\textwidth}
    \centering
    \includegraphics[width=\linewidth]{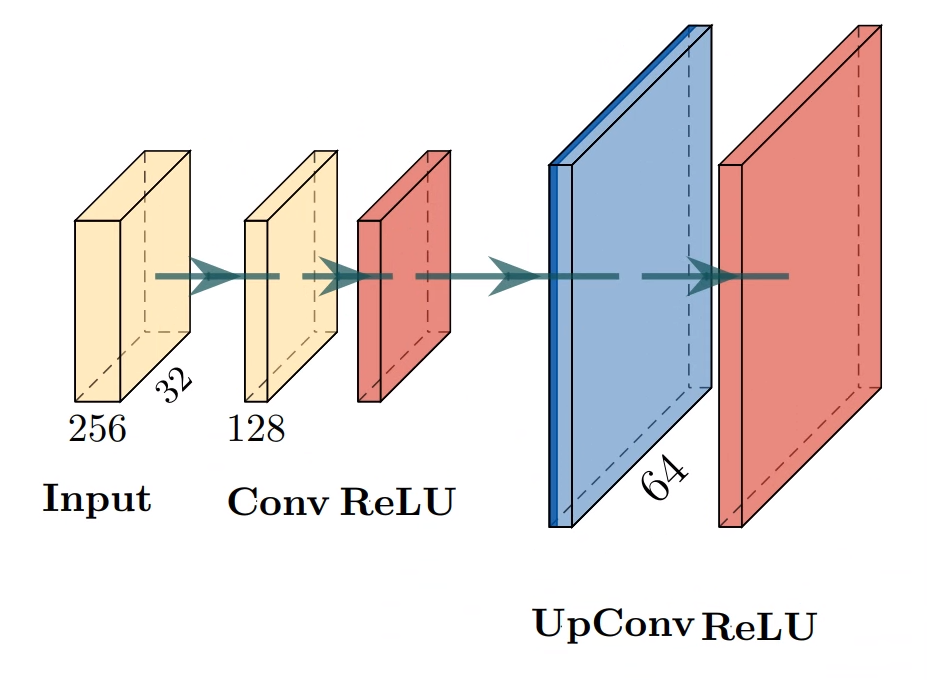} 
    \caption{U-Net decoder block}
\end{subfigure}
\begin{subfigure}[t]{0.66\textwidth}
    \centering
    \includegraphics[width=\linewidth]{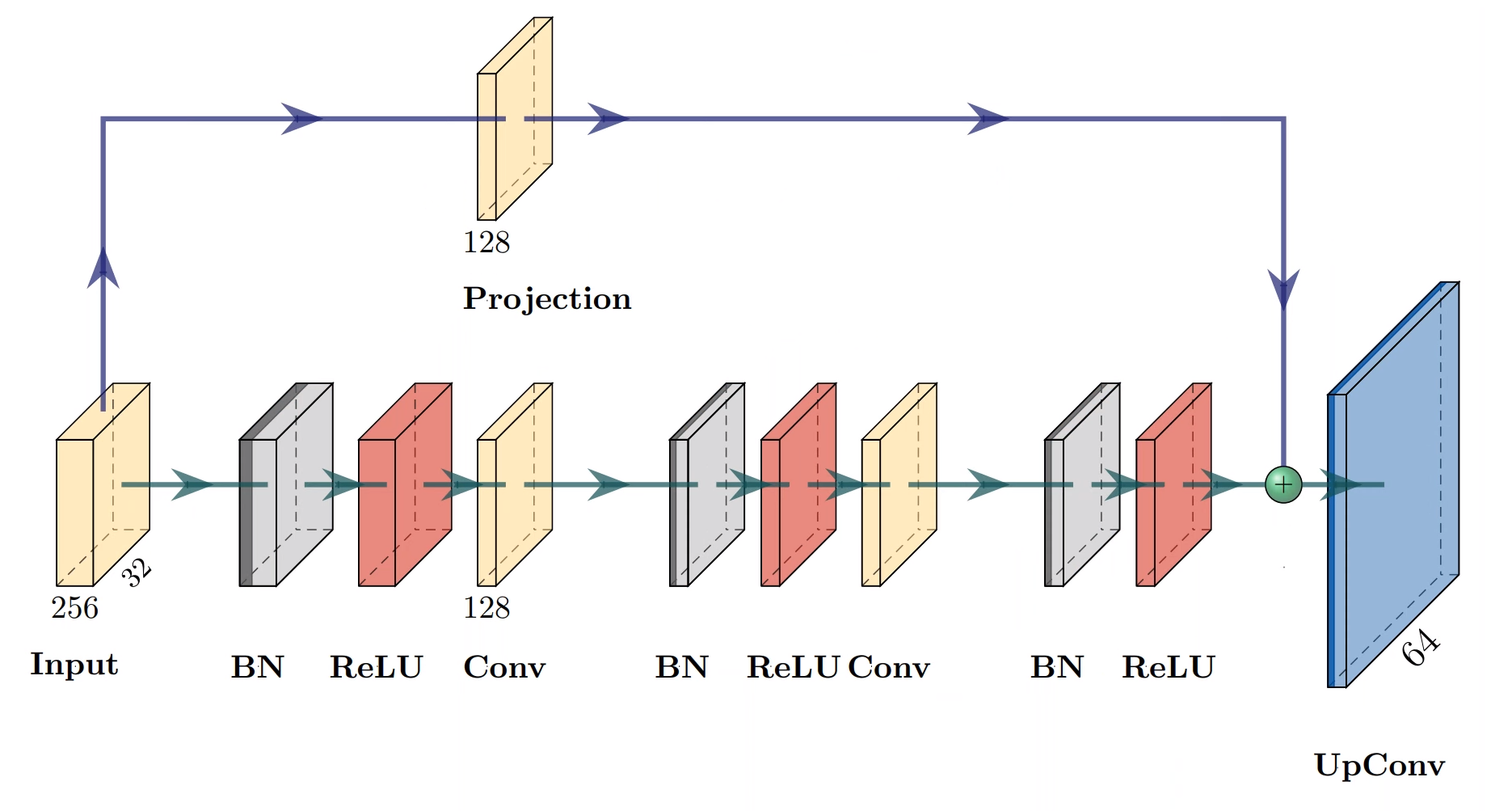} 
    \caption{Residual decoder block}
\end{subfigure}
\caption{Decoder block structures, (a) as defined for U-Net, and (b) a modified version with batch norm and residual connection used in our model.}
\label{fig:decoder}
\end{figure}

U-Net~\cite{unet} and TernausNet-v2~\cite{iglovikov2018ternausnetv2} both employ simple decoder blocks consisting of two (U-Net) or one (TernausNet-v2) convolutional layer(s) and an upconvolution (also known as transposed convolution or deconvolution) for upscaling the feature map by a factor of 2. No normalization is typically performed. One common modification to this structure is simplifying layers even further, e.g. employing bilinear upsampling instead of upcovolution and skipping some of the decoder blocks altogether. Such modification is often employed in the DeepLab~\cite{chen2018:deeplab-EncoderDecoderWA} model family without any significant performance loss. We have found that increasing the decoder complexity can however bring performance gains, at least for the task we consider in this paper. Inspired by ResNet-v2~\cite{resnetv2} residual blocks, we built a decoder block consisting of two (batch normalization, ReLU, convolution) applications followed by another (batch normalization, ReLU), residual connection to the input and finally an upconvolution, as illustrated in Figure~\ref{fig:decoder}. We hypothesize that the need for precise pixel-wise annotations of small objects means that extra decoder complexity is beneficial in this case; buildings can be as small as 6x6 pixels. We cannot rule out other possibilities though, such as mere parameter number increase or batch normalization affecting model performance positively.

\section{Loss functions}
\label{sec:losses}

For each pixel $i$, the model gives a softmax confidence of being in the building class $\hat{y}_i \in [0,1]$, and we have a ground truth label $y_i \in \{0,1\}$. Cross entropy loss is defined as: 
\begin{equation}
    L_\mathrm{CE}(y, \hat{y}) = -\sum_i \omega_i \left[ y_i\log{\hat{y}_i} + (1 - y_i)\log{(1-\hat{y}_i}) \right]\ ,
\label{eq:cross_entropy_loss}
\end{equation}
where $\omega_i$ is a weight controlling the importance of the $i$th pixel, discussed in the next section. 

Previous work on building detection has shown that mixing cross entropy loss with Dice loss is effective \cite{iglovikov2018ternausnetv2}. We observed in informal experiments some further improvement with a closely related formulation, Focal Tversky Loss, which is defined as:
\begin{equation}
    L_\mathrm{FTL}(y, \hat{y}, \beta, \gamma) = \left( 1 - \frac{\sum_i y_i \hat{y}_i + \epsilon}{\sum_i (1-\beta)y_i + \sum_i \beta \hat{y}_i + \epsilon} \right) ^\gamma\ ,
\end{equation}
where $\beta$ is a parameter controlling the trade-off between false positives and false negatives, and $\gamma$ is a focal parameter that changes the relative importance of `easy' ($\hat{y} \approx y$) and `hard' examples. Our overall loss is given by:
\begin{equation}
    L = L_\mathrm{CE} + \alpha L_\mathrm{FTL}\ ,
\label{eq:mixed_loss}
\end{equation}
using parameters $\alpha=0.5$, $\beta=0.99$ and $\gamma=0.25$, and the constant $\epsilon=10^{-6}$ providing numerical stability.

We note that focal losses tend to use $\gamma>1$, which increases the relative importance of difficult examples. In informal experiments we observed, however, that test set performance deteriorated when using $\gamma > 1$. As the optimal setting in our experiments boosted the easy examples, we hypothesise that in our training set, some of the `difficult' examples actually were mislabelled, which was supported by visual inspection of training examples with high loss scores.
The focal parameter in this case helps to make the loss robust to label noise.

\subsection{Weighting}

When all pixels are weighted equally, i.e.\ $\omega_i = 1$ for all $i$ in Eq. (\ref{eq:cross_entropy_loss}), predictions using the above loss are sub-optimal for building detection. As the authors of U-Net have noted \cite{unet}, to distinguish instances it helps to emphasise the weighting of the pixels at the edges of nearby or touching instances. Pixels in background regions which are far from any instance can be down-weighted.

\begin{figure}
    \centering
    \includegraphics[width=4cm]{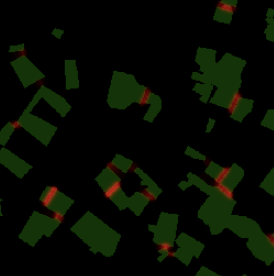}
    \hspace{0.2cm}
    \includegraphics[width=4cm]{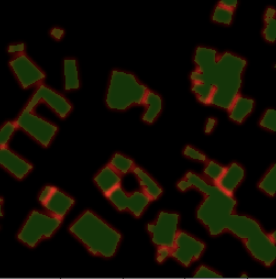}
    \caption{Distance weighting schemes to emphasise nearby edges: U-Net (left) and Gaussian convolution of edges (right). See text for details.}
    \label{fig:distance_weighting}
\end{figure}

The computation for distance-based pixel weighting in \cite{unet} is:
\begin{equation}
    \omega_i = \exp \left( - \frac{ d_1(i) + d_2(i) }{2\sigma^2}\right)\ ,
\end{equation}
where $d_1(i)$ and $d_2(i)$ are the Euclidean distances from pixel $i$ to the closest point of the nearest and second-nearest instance, respectively. Values of this weighting are shown for an example in Figure~\ref{fig:distance_weighting} (left).

We found this formulation to be effective, but slow to compute. The calculation of $d_1(i)$ and $d_2(i)$, involving distance transforms for every instance in an image, is not computationally efficient during training. Using this method it is therefore necessary to pre-compute weights, which limits the possibilities for data augmentation.
Therefore, we use an alternative weighting scheme:

\begin{enumerate}
\item Use the labels $y$ to construct an edge image $E$, where $E(i)$ is set to 1 if the pixel at location $i$ is on the boundary of an instance, and zero otherwise. 
\item The pixel weights $\omega$ are given by convolving $E$ with a Gaussian kernel having length scale $\sigma$, then scaling by a constant $c$.
\end{enumerate}

We used settings of $\sigma=3$, $c=200$.
Example values of this Gaussian convolution method are shown in Figure~\ref{fig:distance_weighting} (right). We found this method to give better final performance in building detection, and to be efficient enough to compute on the fly during training.

\section{Regularization}
\label{sec:regularization}

We use a standard set of image augmentations to provide regularization during training: random crops (to obtain a 448$\times$448 patch from the full 600$\times$600 image), horizontal and vertical flips, rotations, and random modifications to the brightness, hue, saturation, and contrast. We observed that the augmentations to color helped the model to generalise to over- and under-exposed overhead images, as well as images in which visibility was low due to atmospheric conditions.

We also use mixup \cite{zhang2017mixup} as a regularization method, initially proposed as a method for classification and which we modify here for segmentation. During training with this method, a random pair of images $x$ and $x'$ are combined with a weighted average:
\begin{equation}
\tilde{x} = \lambda x + (1 - \lambda) x'\ ,
\end{equation}
where $\lambda$ is the mixup ratio coefficient ($\lambda \in [0, 1)$).

The model then makes a prediction $\hat{y}$ on this averaged image, for which cross entropy loss is computed on both sets of corresponding labels $y$ and $y'$, and then combined:
\begin{equation}
\tilde{L}_\mathrm{CE} = \lambda L_\mathrm{CE}\left(\tilde{x}, y\right) + (1 - \lambda) L_\mathrm{CE}\left(\tilde{x}, y'\right)  .
\end{equation}
Note that in the original mixup specification \cite{zhang2017mixup}, a single loss is computed on averaged labels. However, in preliminary experiments we found this not to work as well due to our use of pixel weighting, and so in this case the labels are not averaged. Note also that we use mixup only for the cross-entropy loss term in Eq. (\ref{eq:mixed_loss}), and do not apply it to Focal Tversky loss. We set $\lambda = 0.05$ in our experiments.

\section{Pre-training}
\label{sec:pretraining}

A common practice is to begin training models with weights initialised from an ImageNet \cite{imagenet} classifier. In the case of the U-Net model, the encoder stages can be initialised in this way; the decoder is then randomly initialised. Attempting to improve on this, we investigated the use of domain-specific pre-training methods, on the grounds that the images in the ImageNet dataset have different characteristics than satellite imagery. The datasets described in Section~\ref{sec:pretrain-datasets} provided tasks with which to pre-train classifier models: the night-time luminance prediction task as proposed by Xie et al. \cite{xie2016transfer}, and the prediction of location in the world at either coarse granularity or fine granularity. We trained a variety of ResNet-50 classifier models using these datasets, and evaluated the performance of the U-Net building detection model when using these classifiers to initialise the encoder weights.

Using the three pre-training tasks on their own gave poor performance in building detection. Informally, we visually inspected the 7$\times$7 root block filters learned in the initial layer of the ResNet models, and observed that many of the values were close to zero. Speculating that this was caused by our satellite image datasets being more homogeneous in appearance than ImageNet, we tried two variations of pre-training. The first was to start with an ImageNet classifier and then fine-tune the full model on each pre-training task. In this case, the ImageNet weights appeared to be close to local optima, as the model weights did not greatly change during this fine-tuning. The second strategy was to co-train, in which we set up ResNet-50 models with two classification heads: ImageNet and \{Luminance | Coarse location | Fine location\}. Training batches in this setup contained a mixture of ImageNet and satellite images, with loss computed for the corresponding head. 

In practice, ImageNet pre-training was an effective strategy, which we ultimately used in our detection model. Fine-tuning with nighttime luminance raised average mAP, though not significantly. A comparison is given in Section~\ref{sec:evaluation}. One issue may have been that the pre-training schemes that we considered were all classification tasks, yet the problem we are ultimately interested in is segmentation. The use of segmentation tasks for pre-training would allow initialisation of the decoder, for instance, which may improve final detection performance and training data efficiency.

\section{Self-training}
\label{sec:selftraining}

\begin{figure}[tbp]
\centering
\begin{subfigure}[t]{0.24\textwidth}
    \centering
    \includegraphics[width=\linewidth]{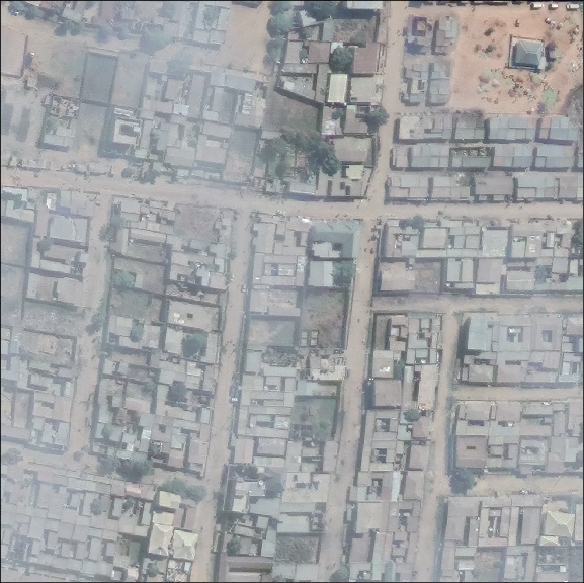} 
    \caption{Input image}
\end{subfigure}
\begin{subfigure}[t]{0.24\textwidth}
    \centering
    \includegraphics[width=\linewidth]{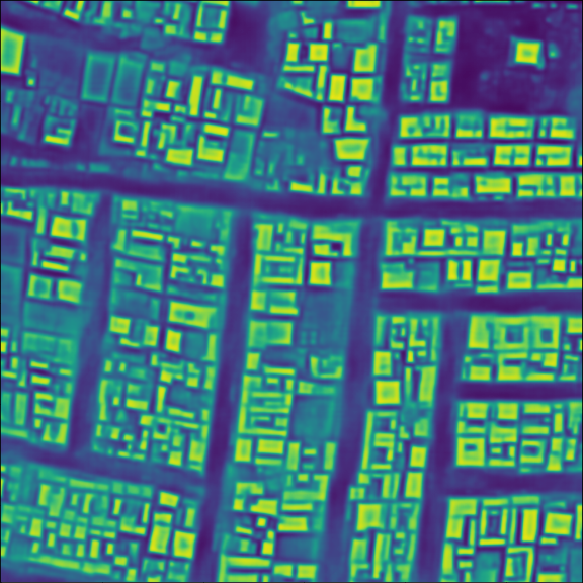} 
    \caption{Teacher confidence}
\end{subfigure}
\begin{subfigure}[t]{0.24\textwidth}
    \centering
    \includegraphics[width=\linewidth]{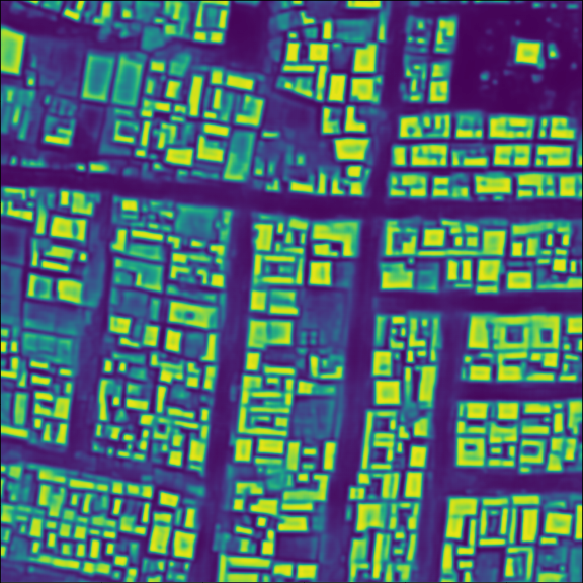} 
    \caption{Student confidence}
\end{subfigure}
\begin{subfigure}[t]{0.24\textwidth}
    \centering
    \includegraphics[width=\linewidth]{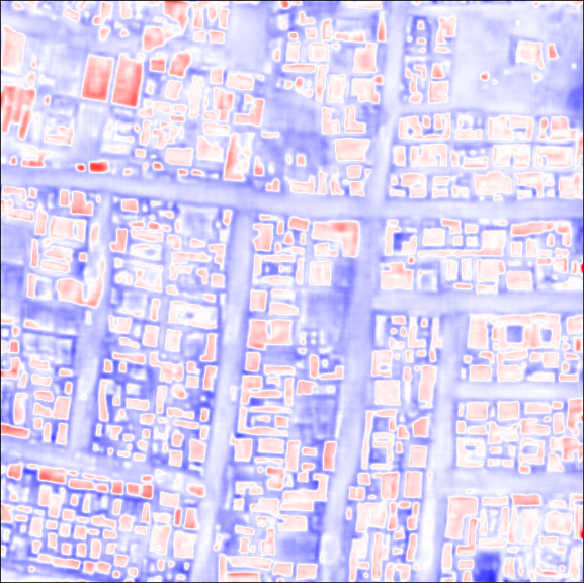} 
    \caption{Difference}
\end{subfigure}
\caption{Comparison of the confidence mask between the teacher and the student after one iteration of self-training. In panel (d), red areas are those that the student model finds more likely to be buildings than the teacher model, and blue areas more likely to be background.}
\label{fig:self_training_improvement}
\end{figure}

In comparison with the limited amount of labeled data, a much larger amount of unlabeled satellite images exists. Leveraging this fact, we employ self-training to improve the model's performance, inspired by the \textit{Noisy Student} \cite{xie2020self} and \textit{Naive Student} \cite{chen2020naive} approaches. For self-training we use the unlabeled dataset described in Section~\ref{subsec:self-training-dataset} and similar image augmentations as for labeled data. See Figure~\ref{fig:self_training_improvement} for a visualization of the performance improvement due to self-training.

We arrived at our best model by performing multiple iterations of self-training using soft teacher labels together with a Kullback-Leibler divergence loss with a focal $\gamma = 0.25$ parameter \cite{focal_kl}. Based on informal experiments using hard teacher labels with the previously defined supervised losses (Section~\ref{sec:losses}), $\gamma \geq 1$, larger students and stochastic depth~\cite{huang2016deep} did not improve performance. Fine-tuning the student model on our original supervised data helped only for the first iteration. 

Some of the satellite images we used had black regions due to extending beyond the satellite image asset geometry, and we observed that our best supervised model (first teacher) failed to detect buildings next to these black pixel parts. It was caused by incorrectly labeled supervised data. We managed to leverage self-training with random black mask augmentation to generate a student model that does not have this issue.

\section{Pre-processing}
\label{sec:preprocessing}

\paragraph{Erosion of instances}

We noticed that in some examples the buildings are so close that the instances effectively touch each other and form one connected component on the segmentation mask. To be able to separate these buildings during post-processing (to identify instances) we had to teach the model to predict at least one pixel gap between them. Therefore we employed a morphological erosion operation with kernel size $3\times3$ pixels during pre-processing of labeled images to shrink all instances by one pixel.

\paragraph{Mapping dense labels}

During training, we remapped dense building labels (representing a group of buildings) to normal building labels.  An alternative is to set the pixel weight to 0 for dense buildings, effectively treating them as `unknown', which was equivalent in terms of performance.

\section{Post-processing}
\label{sec:postprocessing}

\paragraph{Ensembling and test-time augmentation}

To improve the final performance of the model we combined ensembling and simple multi-scale test-time augmentation. In case of ensembling we take an average of the output confidence masks from multiple models on the same input and in case of test-time augmentations we average the confidence masks produced at different image scales (1, $\frac{512}{448}$, and $\frac{576}{448}$).

\paragraph{Connected components}

Our model is a semantic segmentation model, therefore to obtain building instances we find the 4-connected components in the thresholded predicted label image. We calculate the instance confidence score as the average of confidence scores of the connected component.

\paragraph{Dilation of instances}

During pre-processing we applied erosion with kernel size $3\times3$ to instance masks, to shrink them by one pixel. In post-processing we approximate the inverse of this operation by performing morphological dilation on each instance, with the same kernel.

\section{Evaluation}
\label{sec:evaluation}

\begin{table}[tbp]
\caption{U-Net supervised learning baseline configuration.}
\normalsize
\centering
\begin{tabular}{rl}
Encoder:            & ResNet50                                     \\
Decoder block:      & Residual                                      \\
Loss:               & Weighted cross entropy and focal Tversky loss \\
Distance weighting: & Gaussian convolution                          \\
Regularization:     & Image augmentations, mixup                 \\
Pre- and post-processing: & Erosion and dilation
\end{tabular}
\label{tab:supervised-baseline-configuration}
\end{table}

We carried out an ablation study to determine the contribution of the techniques described in the preceding sections. The baseline configuration for supervised learning, with the combination of methods that we found in preliminary experiments to be most effective, is summarised in Table~\ref{tab:supervised-baseline-configuration}. We use the training and test sets as described in Section~\ref{sec:fully-labelled-data}, and train using a scaled conjugate gradient optimizer, with initial learning rate 0.2, decaying by a factor of 0.8 every 10k steps, for a total of 100k steps with batch size 128. Our test set performance metric is mean average precision with an intersection over union threshold of 0.5 (mAP@0.5IoU), using COCO metrics \cite{caesar2018coco}.

\begin{figure}[tbp]
\centering
\includegraphics[width=0.9\linewidth]{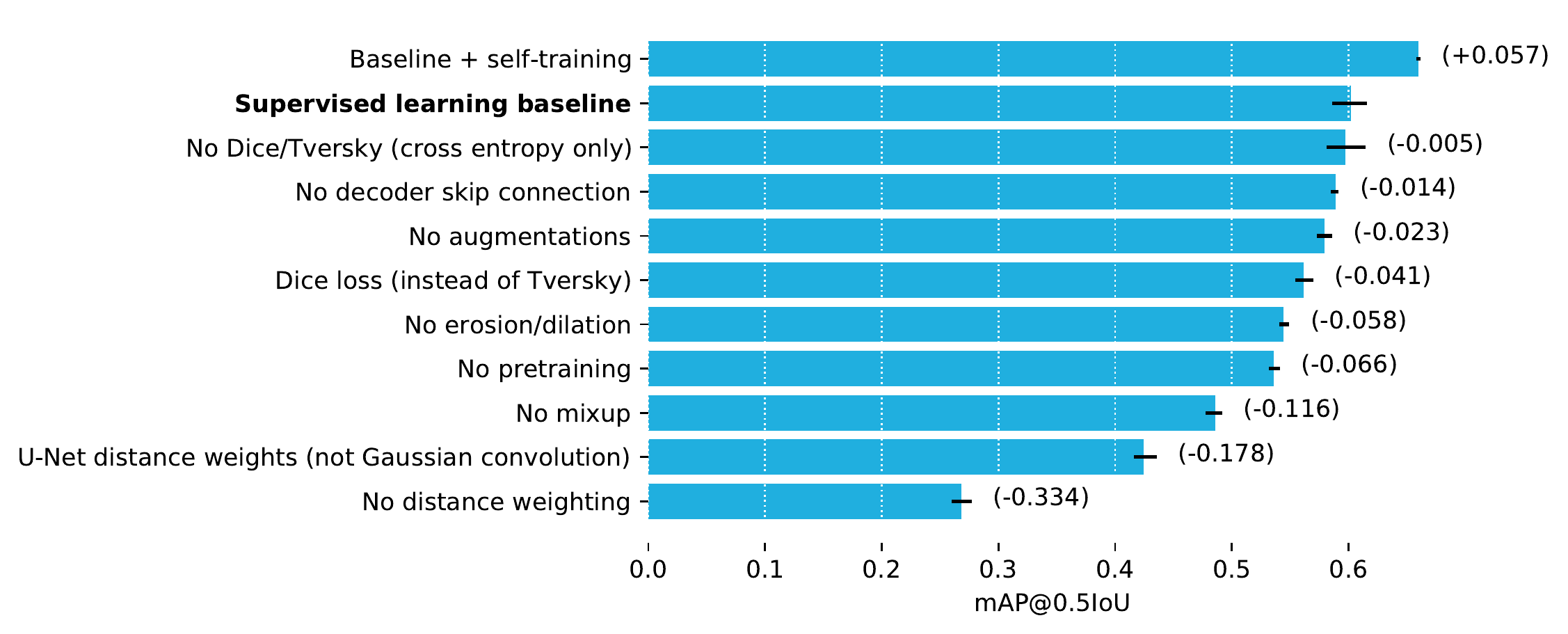} 
\caption{Ablation study of training methods. The first row shows the mAP performance of best model including self-training, and the second row shows the best model with supervised learning only (the baseline). By disabling each training optimisation in turn from the baseline, we observe the impact on mAP test performance: distance weighting has the most significant effect, followed by mixup.}
\label{fig:ablation}
\end{figure}

Ablations were done by changing one configuration setting at a time and measuring the drop in performance relative to the baseline. The results are shown in Figure~\ref{fig:ablation}. The method most significantly contributing to detection performance was distance weighting of pixels in cross entropy loss. Finding the correct boundaries of buildings appears to be the crux of the problem, and distance weighting encourages the model to focus on the correct classification for those pixels. Mixup and ImageNet pre-training were the next most significant methods. One surprising finding from this study was that detection performance using only cross entropy loss was nearly as good as the baseline, with only -0.005 mAP difference (not a statistically significant difference, given the range of variation across replicas).

\paragraph{Self-training} 

\begin{figure}[tbp]
\centering
\includegraphics[width=\linewidth]{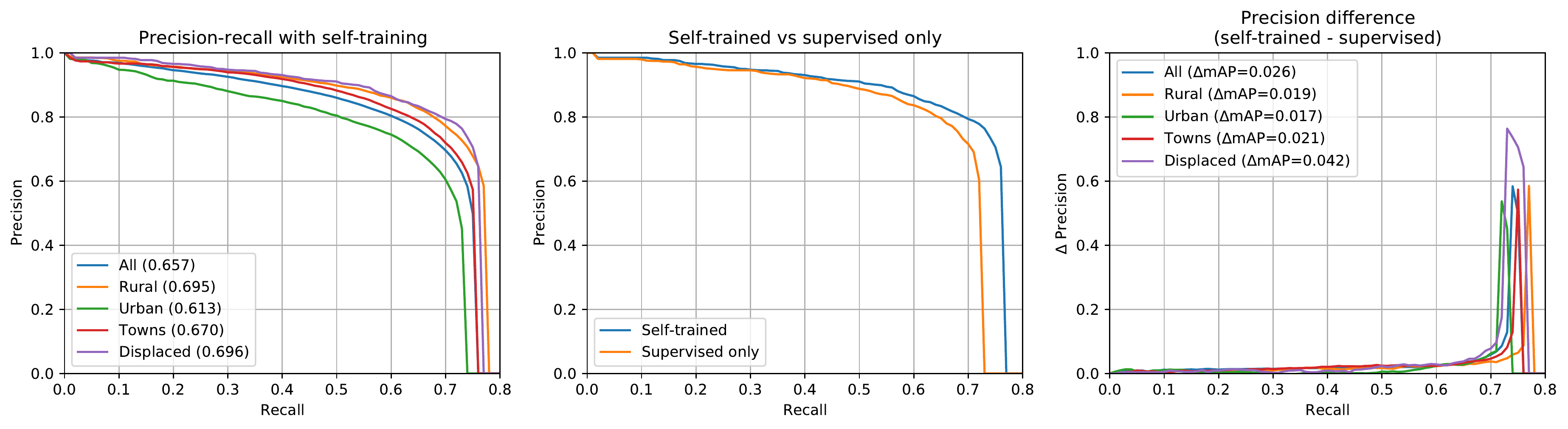} \caption{Precision-recall with IoU threshold 0.5, after self-training. Left: Results on different categories on test images. Centre: overall difference in precision-recall on test data compared to the model before self-training, showing that self-training increases precision at higher recall levels. Right: Difference in precision at each recall level, broken down by different categories of test data.}
\label{fig:pr-curves}
\end{figure}

Our best model was obtained by using the supervised learning baseline as a teacher and carrying out self-training as described in Section~\ref{sec:selftraining}. The bottom row on Figure~\ref{fig:ablation} shows the difference, with mAP increased by 0.057 on average. Figure~\ref{fig:pr-curves} shows precision and recall for different categories in the test set: rural, urban, medium-density (`towns'), and settlement facilities for refugees/internally displaced people (`displaced'). Visual examples of these categories are shown in Figure~\ref{fig:category-examples}. We also show the difference in precision and recall made by the self-training: precision is increased at high recall levels, with the improvement being consistent across all test set categories.

\begin{figure}[tbp]
\centering
\includegraphics[width=\linewidth]{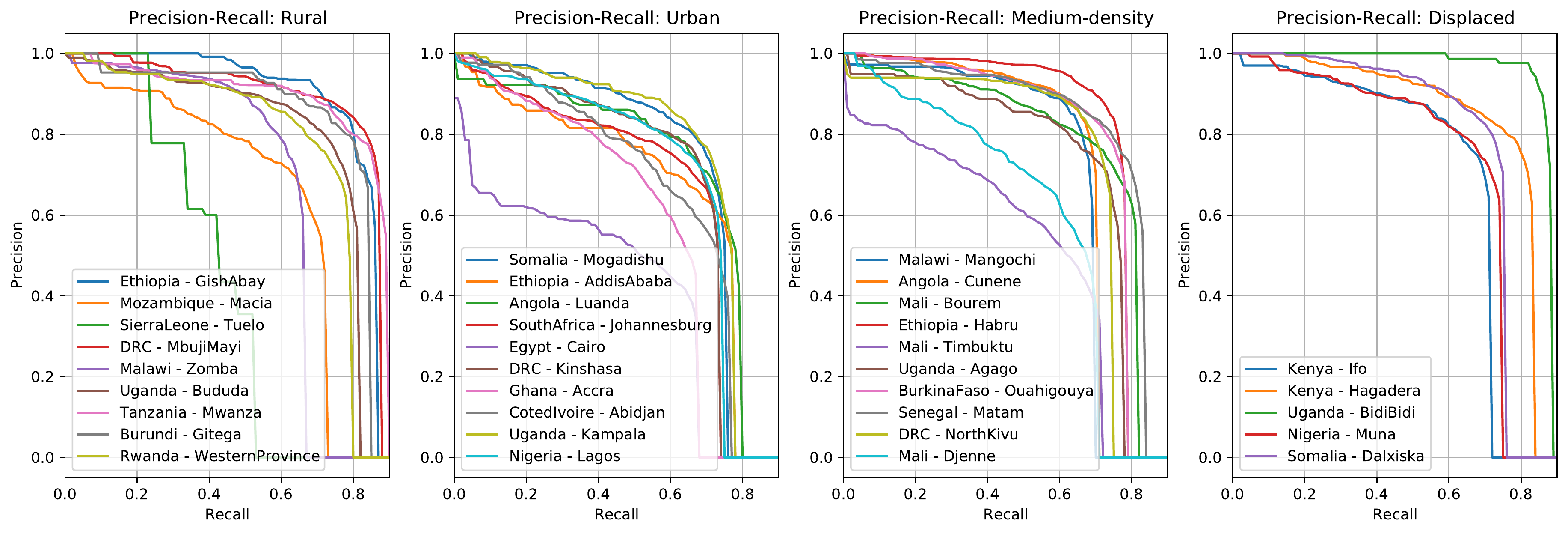} 
\caption{Precision-recall in specific regions by category, of the best model (including self-training). Investigating the regions with low area under the PR curve, we noted that \emph{Sierra Leone - Tuelo} and \emph{Mozambique - Macia} images were sparsely populated, with some buildings missing from the labels (i.e. human error while labelling). \emph{Egypt - Cairo} images were low-scoring partly because of a tendency of the detection model to split large buildings into multiple smaller instances. Detections in desert regions, such as \emph{Mali - Timbuktu} were challenging due to low contrast between roofs and surrounding sandy areas.}
\label{fig:pr-curves-regions}
\end{figure}

\begin{figure}[tbp]
        \centering
        \includegraphics[width=0.8\linewidth]{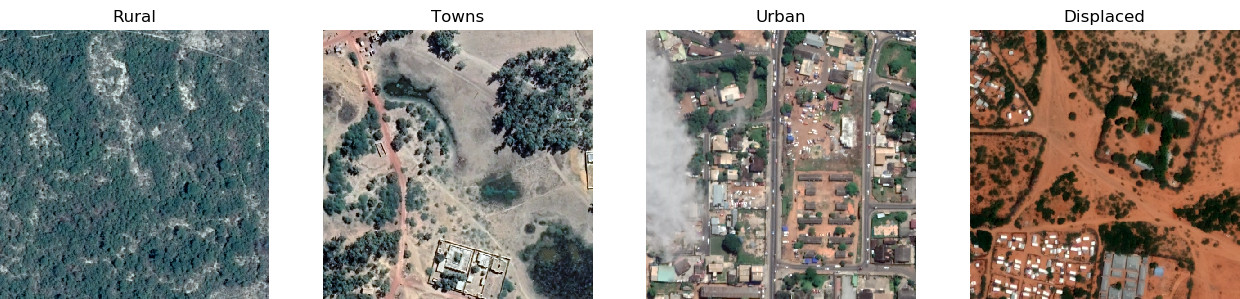} 
        \includegraphics[width=0.8\linewidth]{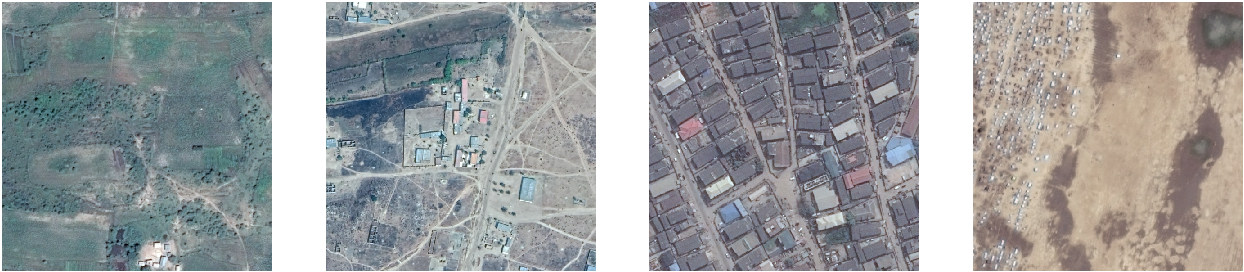} 
        \caption{Examples of the categories evaluated in Figs.~\ref{fig:pr-curves} and \ref{fig:pr-curves-regions}. Imagery: Maxar Technologies.}
\label{fig:category-examples}
\end{figure}

We carried out evaluations of the best model on more specific splits of the evaluation set, shown in Figure~\ref{fig:pr-curves-regions}. When visually inspecting the detections for low-scoring regions, we noted various causes: in rural areas, label errors (single buildings within a mostly-empty area can be difficult for labellers to spot); in urban areas, a tendency of the model to split large buildings into separate instances; and desert terrain, where buildings were hard to distinguish against the background, and the model did not perform as well.

\paragraph{Pre-training}

Table~\ref{tab:pre-training-evaluation} shows the effect of using different weights for initialisation of the encoder. As in the experiments above, we use the supervised learning baseline configuration in Table~\ref{tab:supervised-baseline-configuration}, changing only the initialisation weights. We repeated each experiment five times and report means and confidence intervals. Overall, ImageNet pre-training, optionally with fine tuning based on nighttime luminance, appears to be an effective strategy. 

\begin{table}[tbp]
\caption{Mean average precision of the U-Net building detection model, when using different pre-training schemes to initialise encoder weights.}
\centering
\begin{tabular}{lll}
\toprule
Pre-training scheme & $95$\%  CI mAP \\
\midrule
None & $0.531 \pm 0.003$ \\
ImageNet & $0.601 \pm 0.018$ \\
\midrule
Luminance & $0.579 \pm 0.005$ \\
Coarse location & $0.582 \pm 0.004$ \\
Fine location & $0.583\pm 0.004$ \\
\midrule
\emph{ImageNet, fine tuned with:} & \\
\qquad Luminance & $0.610 \pm 0.006$ \\
\qquad Coarse location & $0.595 \pm 0.004$ \\
\qquad Fine location & $0.602 \pm 0.005$ \\
\midrule
\emph{ImageNet co-trained with:} & \\
\qquad Luminance & $0.552 \pm 0.028$ \\ 
\qquad Coarse location & $0.572 \pm 0.006$ \\
\qquad Fine location & $0.558 \pm 0.019$ \\
\bottomrule
\end{tabular}
\label{tab:pre-training-evaluation}
\end{table}

\begin{figure}[tbp]
\centering
\begin{subfigure}{0.45\textwidth}
    \centering
    \includegraphics[width=\linewidth]{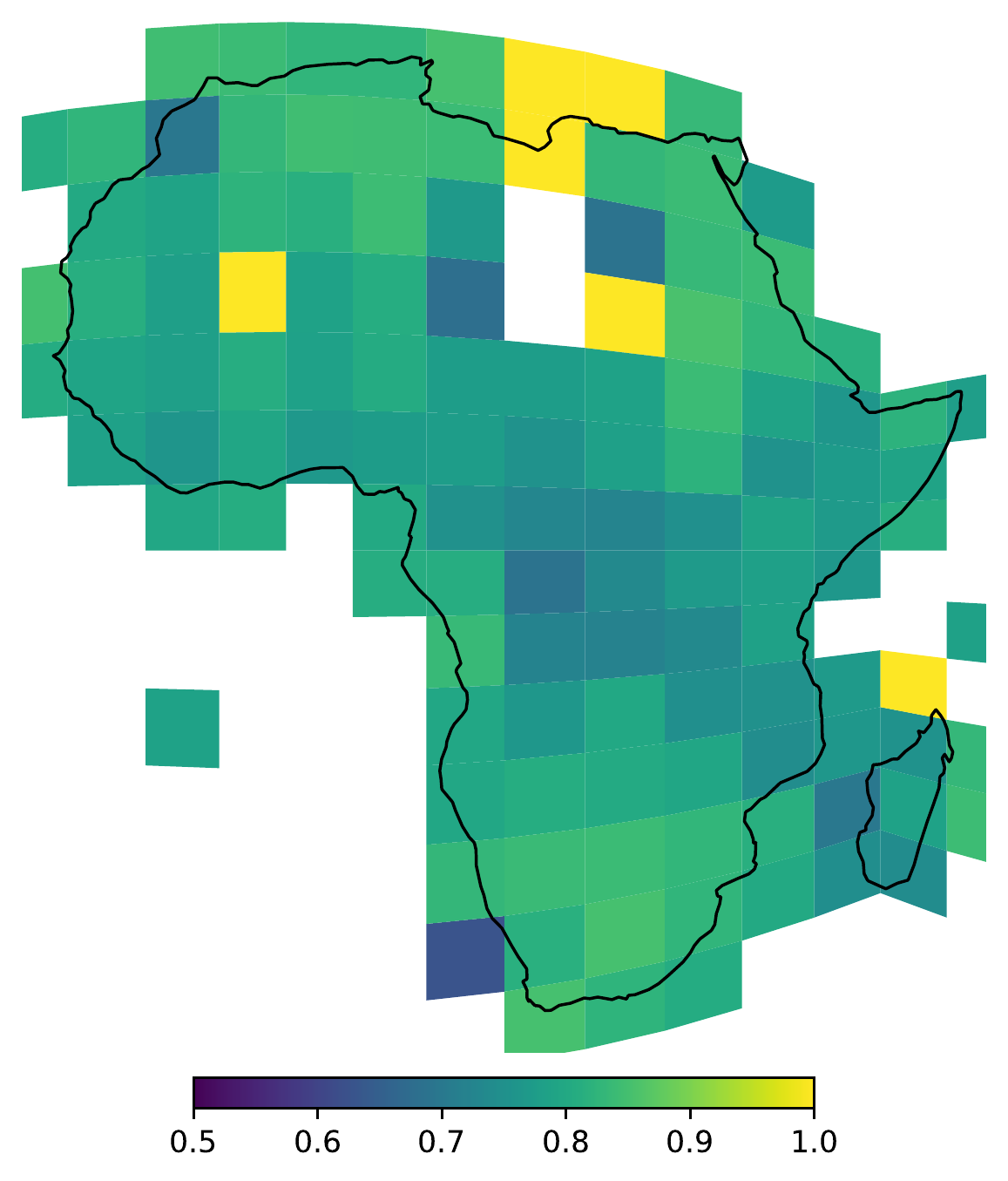} 
    \caption{90\% precision confidence score thresholds.}
\end{subfigure}
\hspace{0.04\textwidth}
\begin{subfigure}{0.45\textwidth}
    \centering
    \includegraphics[width=\linewidth]{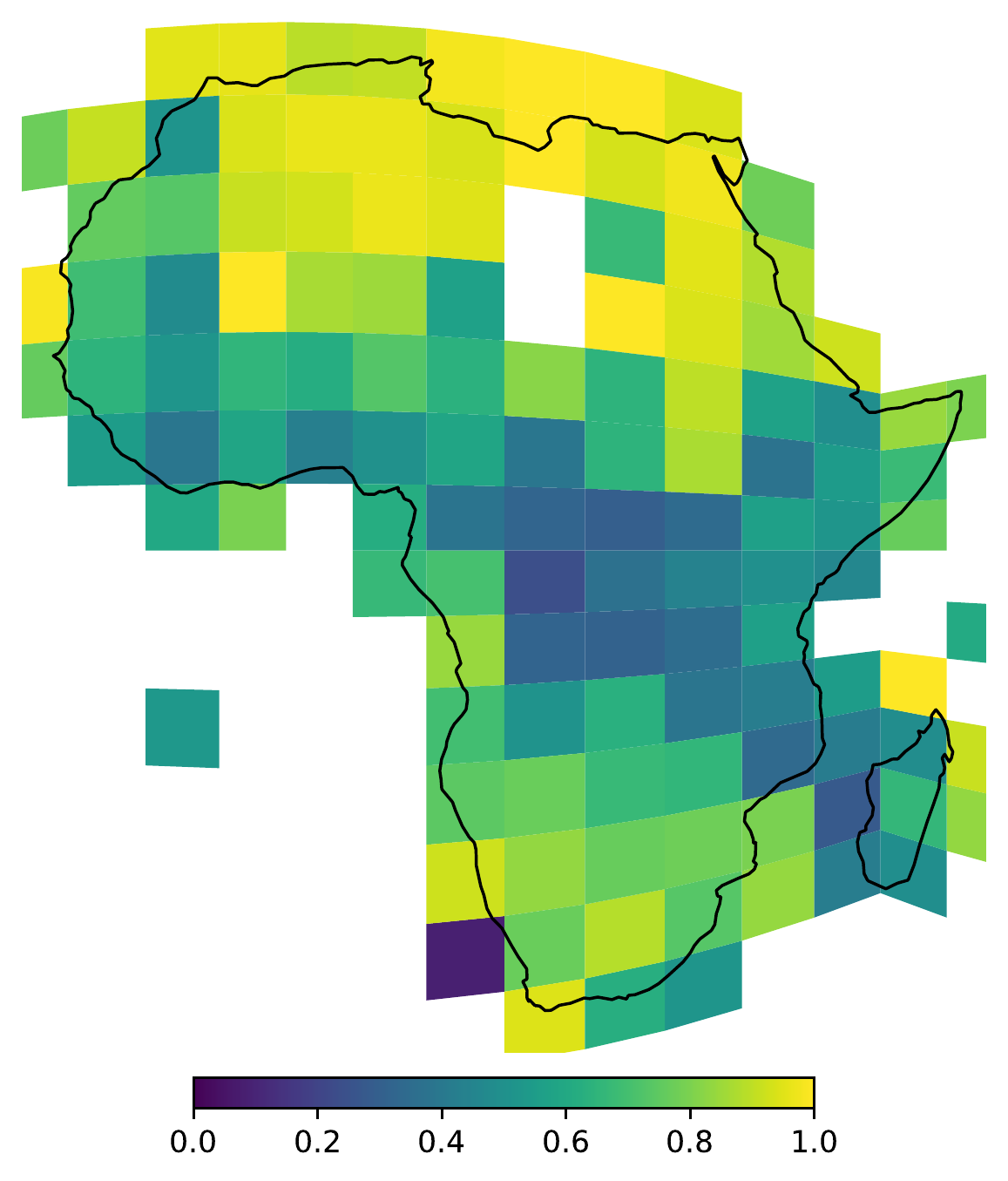} 
    \caption{Fraction of detections dropped at 90\% precision confidence score thresholds.}
\end{subfigure}
\caption{Spatial variations in filtering the full dataset to obtain estimated 90\% precision.}
\label{fig:90_percent_precision_score_threshold_map}
\end{figure}

\begin{figure}
\centering
\begin{subfigure}{0.23\textwidth}
    \centering
    \includegraphics[height=0.13\textheight]{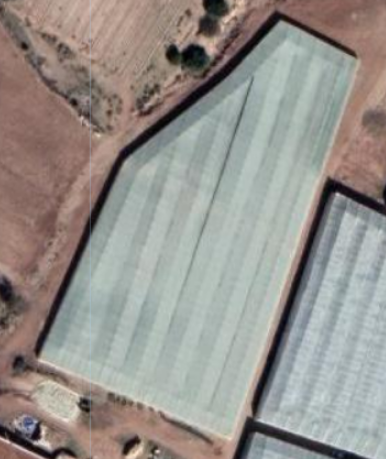}
    \includegraphics[height=0.13\textheight]{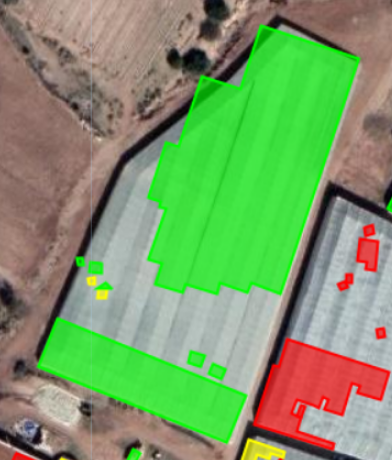}
    \caption{Large buildings}
\end{subfigure}
\begin{subfigure}{0.23\textwidth}
    \centering
    \includegraphics[height=0.13\textheight]{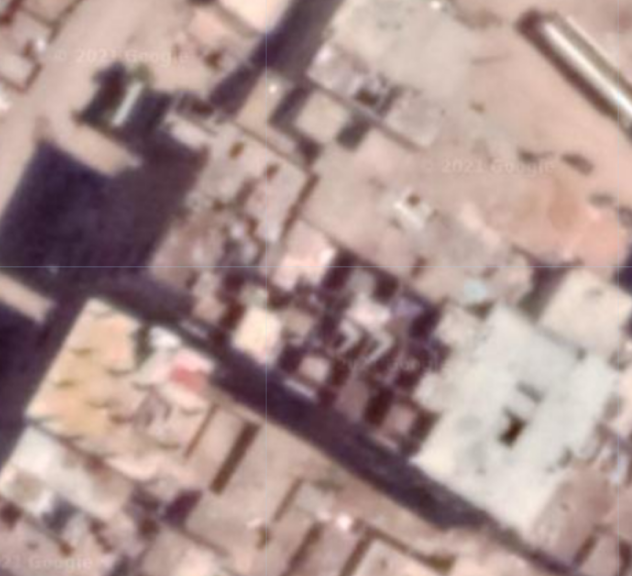}
    \includegraphics[height=0.13\textheight]{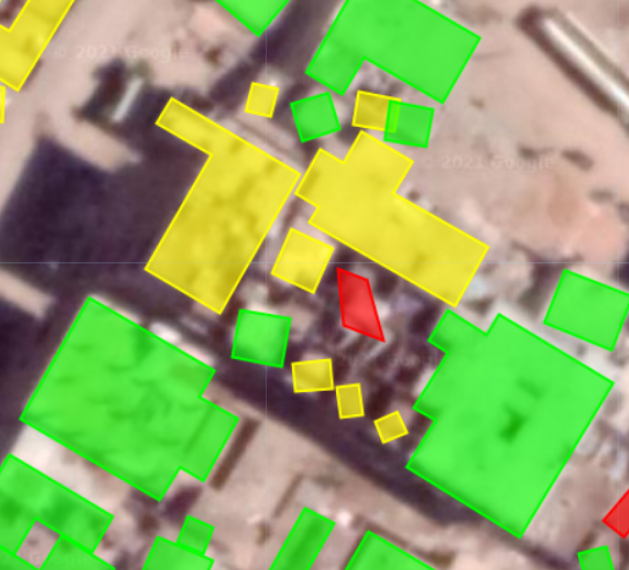}
    \caption{Complex roof structure}
\end{subfigure}
\begin{subfigure}{0.23\textwidth}
    \centering
    \includegraphics[height=0.13\textheight]{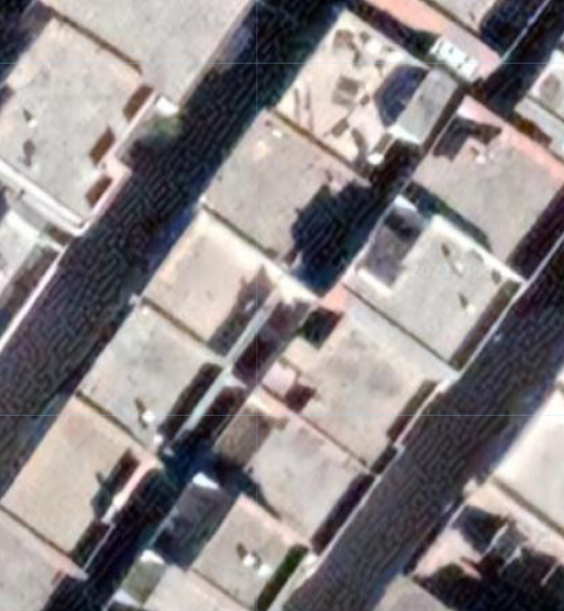}
    \includegraphics[height=0.13\textheight]{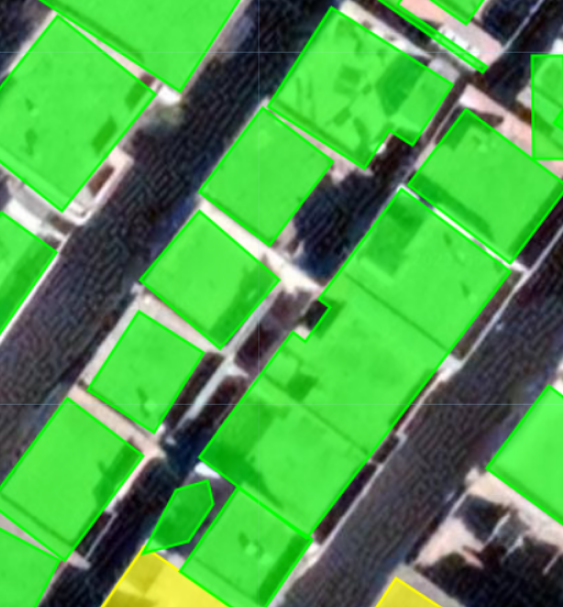}
    \caption{Touching buildings}
\end{subfigure}
\begin{subfigure}{0.23\textwidth}
    \centering
    \includegraphics[height=0.13\textheight]{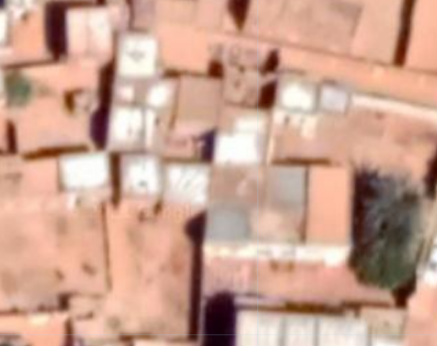}
    \includegraphics[height=0.13\textheight]{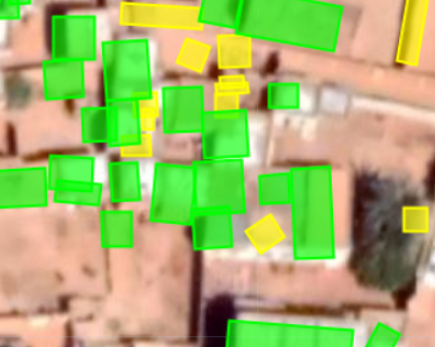}
    \caption{Ambiguous segmentation}
\end{subfigure}

\begin{subfigure}{0.23\textwidth}
\vspace{.3cm}
    \centering
    \includegraphics[height=0.12\textheight]{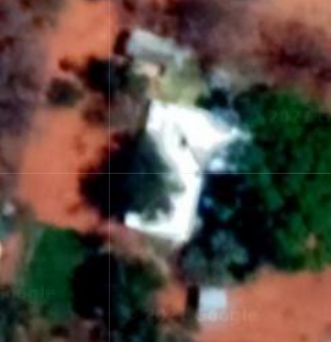}
    \includegraphics[height=0.12\textheight]{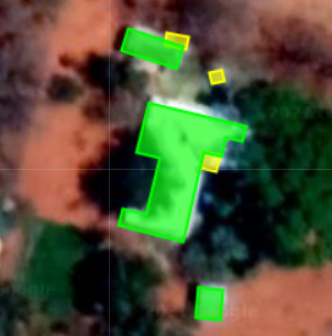}
    \caption{Tree occlusion}
\end{subfigure}
\begin{subfigure}{0.23\textwidth}
\vspace{.3cm}
    \centering
    \includegraphics[height=0.12\textheight]{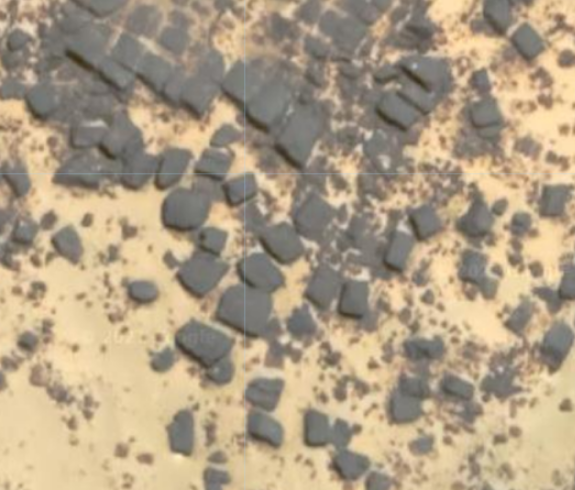}
    \includegraphics[height=0.12\textheight]{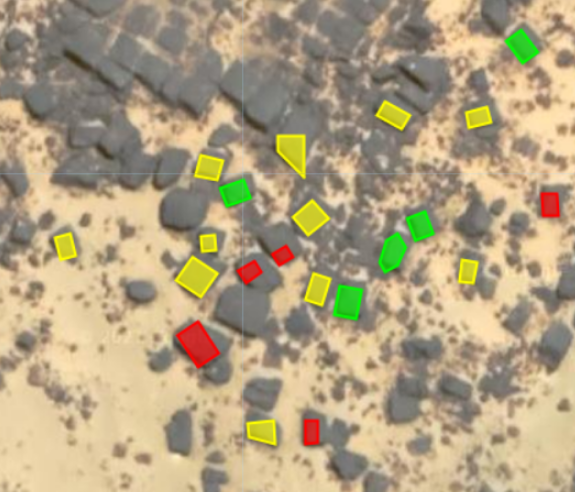}
    \caption{Confusing natural features}
\end{subfigure}
\begin{subfigure}{0.40\textwidth}
\vspace{.3cm}
    \centering
    \includegraphics[height=0.12\textheight]{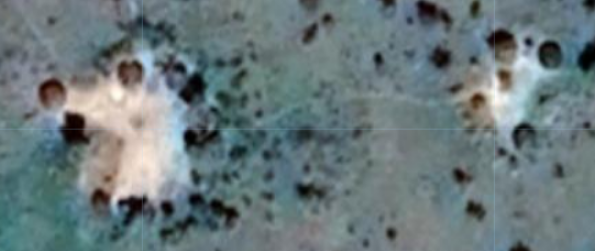}
    \includegraphics[height=0.12\textheight]{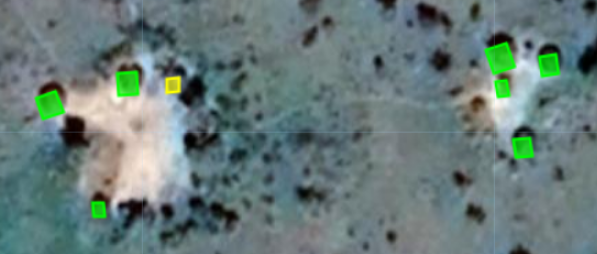}
    \caption{Round shapes vectorized to rectangles}
\end{subfigure}
\caption{Examples of error types occurring in the final dataset, including the contouring and deduplication processes. The color of the polygon indicated the confidence score range: red [0.5;0.6), yellow [0.6;0.7) and green [0.7;1.0]. In panel (d), note also the example of detections appearing shifted, which is caused by misalignment between the image used for inference and the image used for visualization. Orthorectification errors in source imagery can cause building footprints to be generated a few metres from their true positions.}
\label{fig:loss_patterns_final_dataset}
\end{figure}

\begin{figure}[tbp]
\centering
\begin{subfigure}{0.45\textwidth}
    \centering
    \includegraphics[width=\linewidth]{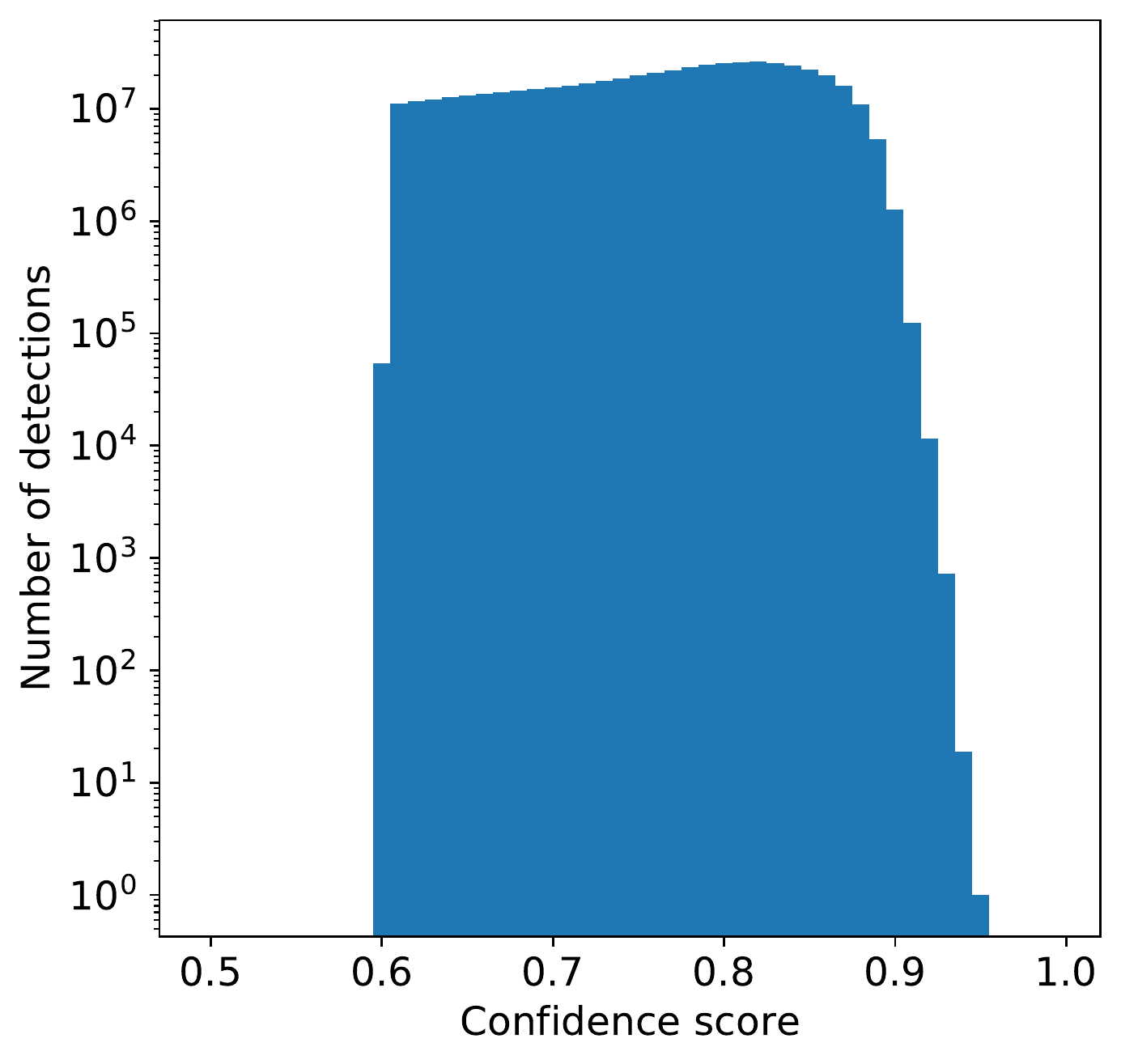} 
\end{subfigure}
\hspace{0.04\textwidth}
\begin{subfigure}{0.45\textwidth}
    \centering
    \includegraphics[width=\linewidth]{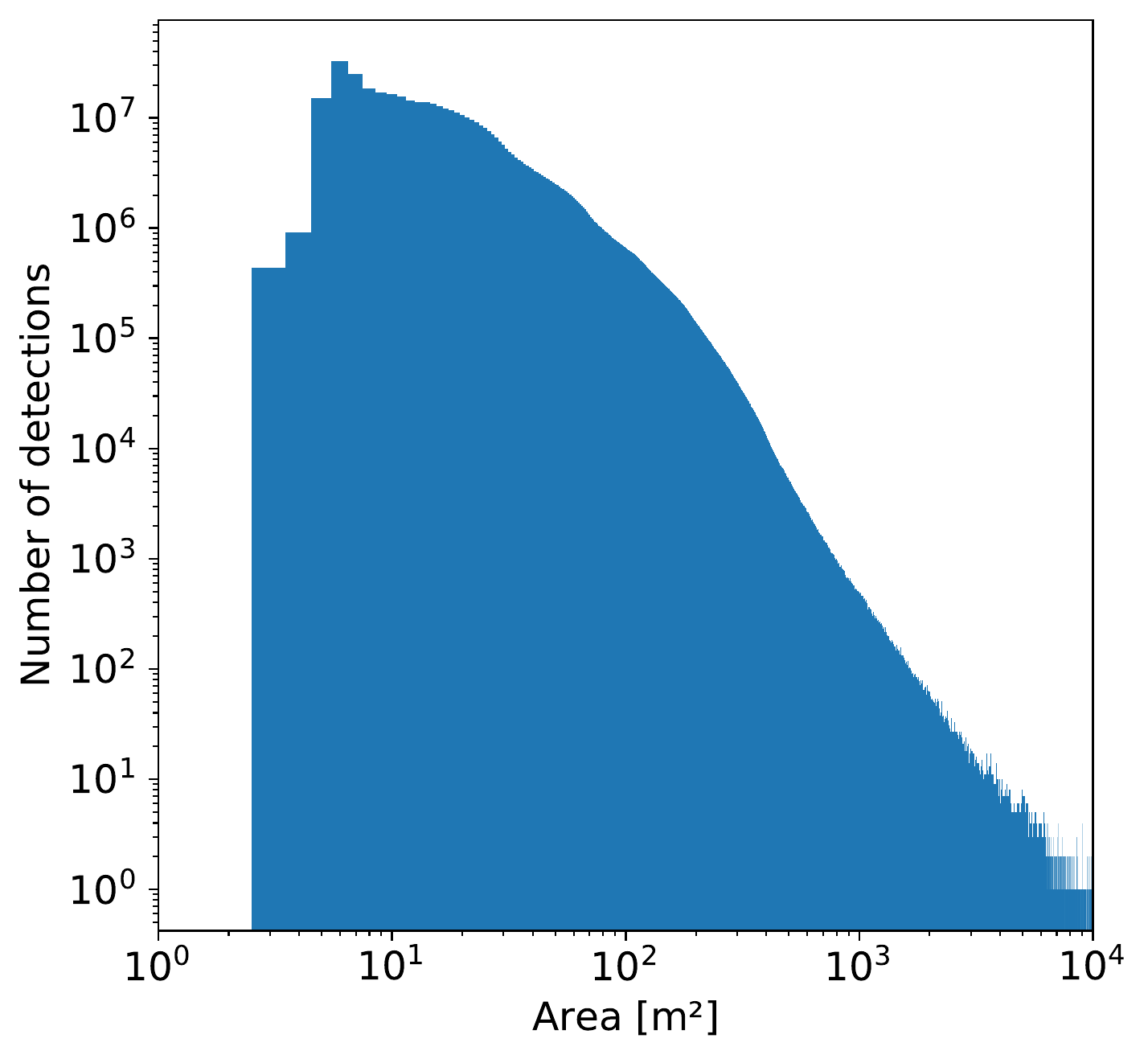} 
\end{subfigure}
\caption{Open Buildings dataset confidence score and area distribution.}
\label{fig:open_buildings_v1_statistics}
\end{figure}

\section{Generation of the Open Buildings dataset}

We used existing infrastructure in Google Maps to run inference, contouring of masks into polygons and deduplication. For inference we used our best model and ran it on available high-resolution satellite imagery in Africa (19.4M km$^2$, 64\% of the land surface of the continent), which included imagery at different timestamps and resolutions. We used a contouring algorithm that produces angular shapes and realigns groups of nearby polygons. After inference and contouring we ended up with 36B building polygons that we deduplicated into 516M polygons (see Figure~\ref{fig:open_buildings_v1_statistics} for statistics). The deduplication algorithm grouped overlapping detections then selected the best polygons based on confidence score of the detections and quality of the imagery. Some potential false positives were removed by the deduplication algorithm if detections on overlapping imagery did not agree.

\paragraph{Confidence score guidelines}

Knowing that model performance varies across regions, we attempted to estimate score threshold guidelines for different regions. These can be used to filter the detections in order to achieve a certain precision level (though with unknown effect on recall). The extra evaluation dataset described in Section~\ref{sec:datasets} provided the means to compute such thresholds for each S2 cell bucket. We reweighted the samples of this extra evaluation data to match the density of the Open Buildings dataset, and then for each level-4 S2 cell, calculated the score thresholds that give 80\%, 85\% and 90\% precision at 0.5 IoU. See Figure~\ref{fig:90_percent_precision_score_threshold_map} for visualization of the 90\% precision score thresholds across Africa. To illustrate the types of errors that cause low precision in the final dataset, Figure~\ref{fig:loss_patterns_final_dataset} shows examples, including model failures on large or complex buildings, and spurious detections in areas with confusing natural features.

\section{Conclusion}

We have presented a pipeline for instance segmentation of buildings in satellite imagery, used to detect buildings across the entire continent of Africa. The methods that we have found for improving detection performance, such as self-training, mixup, and alternative forms of distance weighting, have been applied using the U-Net model, but could in principle be applied to other types of instance segmentation architectures. There are a number of possible directions for improving detection performance further. One is the use of multi-modal imagery, e.g.\ adding Sentinel imagery to the input. Another is the use of detection architectures which explicitly find instances, rather than casting the problem as semantic segmentation. As high-resolution overhead imagery becomes more widely available, improved methods for mapping the built environment can help to make progress on a number of practical and scientific applications. 

\section{Acknowledgements}

We would like to thank several people who helped to make this work possible: Abdoulaye Diack assisted with coordination, Brian Shucker, Rob Litzke, Yan Mayster, Michelina Pallone, Stephen Albro, and Matt Manolides provided advice and assistance with the infrastructure used to create the dataset, Andrea Frome and Mohammad Nassar assisted with preliminary work exploring the use of DeepLab as an alternative basis for detection, Nyalleng Moorosi helped with diligence on ethical and safety issues, and Sean Askay helped to scope the dataset and identify practical applications. The work is part of Google's ongoing AI for Social Good initiative.

\bibliographystyle{unsrt}  
\bibliography{main}  

\end{document}